\documentclass[screen, authorversion, acmsmall]{acmart}
\usepackage{amsmath}
\usepackage{graphicx}
\usepackage{hyperref}
\usepackage[utf8]{inputenc}
\usepackage{natbib}
\usepackage{xcolor}
\usepackage{ifthen}
\usepackage[normalem]{ulem}
\usepackage{svg}
\usepackage{subfiles}
\usepackage{xspace}
\usepackage{cleveref}
\usepackage{wrapfig}
\usepackage{dialogue}
\usepackage[most]{tcolorbox}
\usepackage{bbm}
\usepackage{xspace}
\usepackage{enumitem}
\setlist[itemize]{align=parleft,left=0pt..1em,noitemsep,topsep=0pt}
\usepackage{authblk}
\usepackage[symbol]{footmisc}

\makeatletter
\renewcommand\AB@affilsepx{, \protect\Affilfont}
\makeatother

\newcommand\R{\ensuremath{\mathbb{R}}} 

\ifthenelse{\isundefined{\definition}}{\newtheorem{definition}{Definition}}{}
\ifthenelse{\isundefined{\assumption}}{\newtheorem{assumption}{Assumption}}{}
\ifthenelse{\isundefined{\hypothesis}}{\newtheorem{hypothesis}{Hypothesis}}{}
\ifthenelse{\isundefined{\proposition}}{\newtheorem{proposition}{Proposition}}{}
\ifthenelse{\isundefined{\theorem}}{\newtheorem{theorem}{Theorem}}{}
\ifthenelse{\isundefined{\lemma}}{\newtheorem{lemma}{Lemma}}{}
\ifthenelse{\isundefined{\corollary}}{\newtheorem{corollary}{Corollary}}{}
\ifthenelse{\isundefined{\alg}}{\newtheorem{alg}{Algorithm}}{}
\ifthenelse{\isundefined{\example}}{\newtheorem{example}{Example}}{}
\newcommand{\E}{\ensuremath{\mathbb{E}}} 

\def\sqrtexplained#1{%
  \begingroup
    \sbox0{$#1$}
    \def\underbrace##1_##2{##1}
    \sbox2{$#1$}
    \dimen0=\wd0 \advance\dimen0-\wd2
    \mathrlap{\sqrt{\phantom{\displaystyle#1}\kern\dimen0 }}
    \hphantom{\sqrt{\vphantom{\displaystyle#1}}}
  \endgroup
  #1}

\def\mdp{\mathcal{M}}
\def\pitarget{\pi^*}

\def\Sset{S}
\def\Aset{A}

\def\Dset{\mathcal{D}}
\def\Demos{\mathcal{D}_\mathrm{RL}^*}
\def\Doff{\mathcal{D}_\mathrm{RL}}

\def\Reward{\mathcal{R}}
\def\Return{R}
\def\Oset{\mathcal{O}}
\def\Emit{\mathcal{E}}
\def\Trans{\mathcal{T}}
\def\init{\mu}
\def\defeq{:=}

\def\jrl{J}

\def\dkl{D_\mathrm{KL}}

\usepackage{authblk}

\makeatletter         
\renewcommand\maketitle{
{\raggedright 
\begin{center}
{\Huge \bfseries \sffamily \@title }\\[2ex]
{\@author}\\[2ex] 
\end{center}}}
\makeatother

\renewenvironment{abstract}{
    \newline
    \itshape
    }
{}

\begin{document}
\title{Foundation Models for Decision Making: Problems, Methods, and Opportunities}
\author[1,2]{\mbox{Sherry Yang}\footnote{Corresponding author: sherryy@berkeley.edu}}
\author[1]{\mbox{Ofir Nachum}}
\author[3]{\mbox{Yilun Du}}
\author[1]{\mbox{Jason Wei}}
\author[2]{\mbox{Pieter Abbeel}}
\author[1,4]{\mbox{Dale Schuurmans}}
\affil[1]{Google Research, Brain Team}
\affil[2]{UC Berkeley}
\affil[3]{MIT}
\affil[4]{University of Alberta}

\maketitle
\noindent 
\vspace{-0.2in}
\begin{abstract}
Foundation models pretrained on diverse data at scale 
have demonstrated extraordinary capabilities in a wide range of 
vision and language tasks. 
When such models are deployed in real world environments, they inevitably interface with other entities and agents. For example, language models are often used to interact with human beings through dialogue, and visual perception models are used to autonomously navigate neighborhood streets. 
In response to these developments, new paradigms are emerging for training foundation models to interact with other agents and perform long-term reasoning. 
These paradigms leverage
the existence of ever-larger datasets 
curated for 
multimodal, multitask, and generalist interaction. 
Research at the intersection of foundation models and decision making holds tremendous promise for creating 
powerful new systems that can interact effectively across a diverse range of applications such as dialogue, autonomous driving, healthcare, education, and robotics. 
In this manuscript, we examine the scope of foundation models for decision making,
and provide conceptual tools and technical background for understanding the problem space and exploring new research directions. We review recent approaches that ground foundation models in practical decision making applications 
through a variety of methods such as prompting, conditional generative modeling, planning, optimal control, and reinforcement learning, and discuss common challenges and open problems in the field. 

\begin{figure}[h!]
    \centering
    \includegraphics[width=\textwidth]{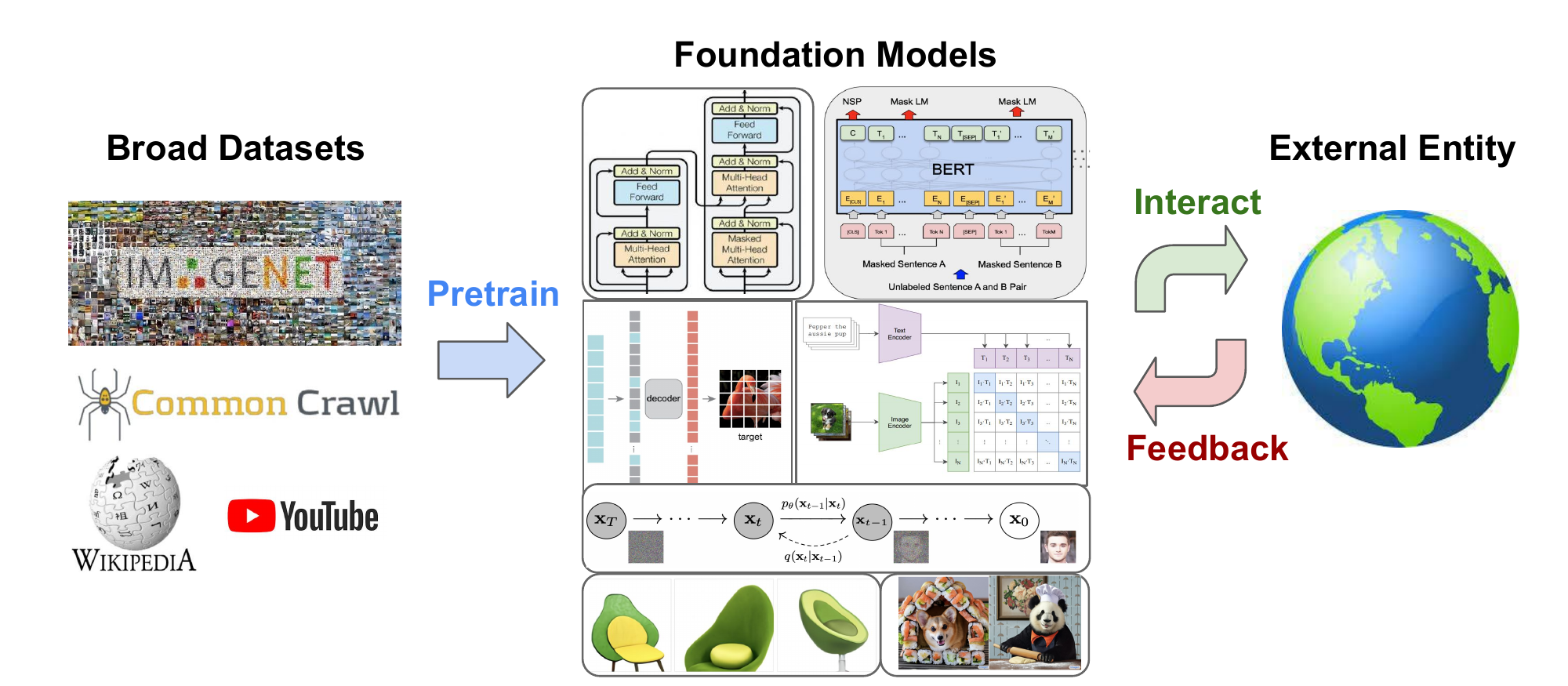}
    \caption{Overview of foundation models for decision making. Foundation models pretrained on broad data are adapted to accomplish specific tasks by interacting with external entities and receiving feedback.}
    \label{fig:fmdm}
\end{figure}

\end{abstract}

\clearpage
\tableofcontents
\clearpage
\hypertarget{introduction}{\section{Introduction}}
\label{sec:introduction}
Foundation models pretrained on broad datasets via self-supervised learning have demonstrated exceptional abilities in knowledge transfer to diverse downstream tasks~\citep{bommasani2021opportunities}. As such models continue to be applied to more complex problems that involve long-term reasoning~\citep{wei2022emergent}, control~\citep{brohan2022rt}, search~\citep{strohman2005indri}, and planning~\citep{huang2022inner}, or are deployed in applications such as dialogue, autonomous driving, healthcare, and robotics, they are expected to interface with external entities and agents. For example, in  dialogue  a language model converses with a human over multiple turns; in robotics a perception-control model executes actions in a real-world environment. These scenarios present new challenges for foundation models, including (1) how to learn from feedback given by an external entity (e.g., human rating of conversation quality), (2) how to adapt to modalities not commonly covered by large language or vision datasets (e.g., robot actions), and (3) how to perform long-term reasoning and planning over the future. 

Such questions have traditionally been at the core of sequential decision making~\citep{sutton2018reinforcement}, encompassing areas such as reinforcement learning, imitation learning, planning, search, and optimal control. Contrary to the paradigm of foundation models, where broad datasets with billions of images and text tokens are used during pretraining, prior work on sequential decision making has largely focused on task-specific or \emph{tabula rasa} settings with limited prior knowledge~\citep{silver2017mastering}. Despite a seemingly disadvantageous setup, research in sequential decision making has achieved significant progress in surpassing human performance on tasks such as playing board games~\cite{tesauro1994td} and Atari video games~\citep{mnih2013playing}, as well as operating robots to complete navigation~\citep{pomerleau1988alvinn} and manipulation tasks~\citep{kalashnikov2018scalable,akkaya2019solving}. 
Nevertheless, since these methods learn to solve a task from scratch without broad knowledge from vision, language, or other datasets, they generally struggle with generalization and sample efficiency, e.g., requiring 7 GPU days of interactive game-play to solve a single Atari game~\citep{agarwal2022beyond}.
Intuitively, broad datasets similar to those used for foundation models should also be beneficial for sequential decision making models. For example, there are countless articles and videos on the Internet about how to play Atari games. Similarly, there is a wealth of knowledge about properties of objects and scenes that would be useful to a robot, or about human wants and emotions that could improve a dialogue model.

While research on foundation models and sequential decision making has largely been disjoint due to distinct applications and foci, 
there is increasing activity at the intersection of these communities.
On the foundation models side, with the discovery of emergent properties of large language models, 
target applications
have graduated from simple zero or few-shot vision and language tasks to  problems that now involve long-term reasoning~\citep{srivastava2022beyond,wei2022chain,lewkowycz2022solving} or multiple interactions~\citep{chatgpt2022}. Conversely, in the sequential decision making communities, researchers inspired by the success of 
large scale vision and language models have begun to curate ever-larger datasets for learning multimodel, multitask, and generalist interactive agents~\citep{agarwal2020optimistic,szot2021habitat,fan2022minedojo,brohan2022rt,reed2022generalist,lee2022multi}. 
Further blurring the lines between the two fields, some recent work has investigated the use of pretrained foundation models such as CLIP~\citep{radford2021learning} and ViT~\citep{dosovitskiy2020image} to bootstrap the training of interactive agents for visual environments~\citep{khandelwal2022simple,tao2022evaluating}, while other work has investigated foundation models as dialogue agents optimized by reinforcement learning with human feedback~\citep{ouyang2022training}, and other work has adapted large language models  to interact with external tools such as search engines~\citep{komeili2021internet,thoppilan2022lamda, lazaridou2022internet,shuster2022blenderbot,yao2022react}, calculators~\citep{cobbe2021training,thoppilan2022lamda}, translators~\citep{thoppilan2022lamda}, MuJoCo simulators~\citep{liu2022mind}, and program interpreters~\citep{gao2022pal}.

Our premise in this report is that research on foundation models and interactive decision making can be mutually beneficial if  considered jointly. On one hand, adaptation of foundation models to tasks that involve external entities can benefit from incorporating feedback interactively and performing long-term planning. On the other hand, sequential decision making can leverage world knowledge from foundation models to solve tasks faster and generalize better. With the aim of spurring further research at the intersection of these two fields, we 
scope the problem space of \emph{foundation models for decision making}. 
We provide technical tools for understanding current research in the space, review remaining challenges and open problems, and speculate on potential solutions and promising approaches to overcome these challenges.

\subsection{Structure of This Report}
This report is divided into 5 major sections. In Section \ref{sec:background}, we review the relevant background and notations of sequential decision making, and present a few example scenarios where foundation models and decision making are better considered jointly. The subsequent three sections are organized around how foundation models can characterize different components of a decision making system. In Section \ref{sec:technical-generation}, we discuss how foundation models can serve as generative models of \emph{behavior} (e.g., skill discovery) and generative models of the \emph{environment} (e.g., for conducting model-based rollouts). In Section \ref{sec:technical-compression}, we discuss how foundation models can serve as representation learners of states, actions, rewards, and transition dynamics (e.g., plug-and-play vision-language models, model-based representation learning). In Section \ref{sec:technical-lm}, we discuss how language foundation models can serve as interactive agents and environments, enabling new problems and applications to be considered under a sequential decision making framework (language model reasoning, dialogue, tool use). Finally in Section \ref{sec:challenge}, we outline open problems and challenges, and propose potential solutions (e.g., how to leverage broad data, how to structure environments, and what aspects of foundation models and decision making can be improved).

\hypertarget{background}{\section{Preliminaries}}
\label{sec:background}

In this section, we review  relevant background on sequential decision making, and present example scenarios to illustrate when and why it is better to consider foundation models and decision making jointly.

\subsection{Sequential Decision Making Preliminaries}
Unlike vision and language domains, where a foundation model is usually trained (and adapted) only once, sequential decision making focuses on learning from \emph{interactive} experience.
We outline this formalism and introduce common algorithms for sequential decision making.

\subsubsection{Interacting with an Environment}
Sequential decision making problems are most often formalized in terms of a Markov decision process (MDP)~\citep{puterman1994markov}, which is defined as a tuple $\mdp\defeq \langle\Sset, \Aset, \Reward, \Trans, \init, \gamma\rangle$ consisting of a state space $\Sset$, an action space $\Aset$, a reward function $\Reward:\Sset\times\Aset\to\Delta(\R)$,\footnote{$\Delta(\mathcal{X})$ denotes the simplex over a set $\mathcal{X}$.} a transition function $\Trans:\Sset\times\Aset\to\Delta(\Sset)$, an initial state distribution $\mu\in\Delta(\Sset)$, and a discount factor $\gamma \in [0, 1)$. A policy $\pi:\Sset\to\Delta(\Aset)$ interacts with the environment starting at an initial state $s_0 \sim \init$. At each timestep $t \ge 0$, an action $a_t\sim\pi(s_t)$ is sampled and applied to the environment, after which the environment transitions into the next state $s_{t+1}\sim\Trans(s_t,a_t)$ while producing a scalar reward $r_t\sim\Reward(s_t,a_t)$.\footnote{We will focus on fully observable MDPs in this article, though an MDP can be extended to a partially observable MDP (POMDP) by introducing an observation space $\Oset$, an emission function $\Emit:\Sset\to\Oset$, and the restriction that policies  can only depend on observations and previous actions.} 

After $\pi$ interacts with $\mdp$ for $H$ timesteps ($H$ can be infinite), an episode (trajectory) is produced $\tau:=\{(s_0,a_0,r_0),(s_1,a_1,r_1),\dots,(s_H, a_H, r_H)\}$. We use $\tau_{t}$ to denote the tuple $(s_t, a_t, r_t)$,  $\tau_{<t}$ to denote a sub-episode up to timestep $t$, $\tau_{\ge t}$ to denote a sub-episode starting from timestep $t$ and ending at $H$, $\tau_{t:t+h}$ to denote a sub-episode from timestep $t$ to $t+h$, and $\tau_s$ or $\tau_a$ to denote only the state or action portion of a trajectory. The return associated with episode $\tau$ is defined as the total discounted sum of rewards $\Return(\tau) := \sum_{t=0}^H\gamma^t r_t$. The \emph{trajectory distribution} of a policy $p_\pi(\tau)$ is determined by 
\begin{equation}
    p_\pi(\tau) = \init(s_0)\Pi_{t=0}^H\pi(a_t|s_t)\Reward(s_t,a_t)\Trans(s_{t+1}|s_t,a_t).
\end{equation}
Trajectories generated by one or multiple policies can be collected in an offline dataset 
$\Doff=\{\tau\}$. We distinguish $\Doff$ from a typical vision or language dataset $\Dset$; $\tau\sim\Doff$ is an \emph{interactive} trajectory involving actions and rewards whereas $x\sim\Dset$ is a \emph{static} image or a text sequence. Nevertheless, foundation model techniques developed for $\Dset$ can also be apply to $\Doff$.

\subsubsection{Imitation Learning}
 In  standard imitation learning, $\Reward$, $\Trans$, and $\mu$ are unknown to the agent. Learning solely takes place from a fixed dataset of demonstrations $\Doff^*=\{(s,a)\}$ previously collected by an expert policy $\pi^*$ interacting with $\mdp$ through $a\sim\pitarget(s)$. The goal of imitation learning is to train $\pi$ on $\Doff^*$ so that $\pi$ closely approximates $\pitarget$ according to some metric, such as the Kullback–Leibler (KL) divergence between the trajectory distributions $\dkl(p_{\pitarget}(\tau)\|p_\pi(\tau))$.
 
\paragraph{Behavioral cloning (BC).} Learning from expert demonstrations leads to the common framing of imitation learning as supervised learning of state to action mappings. Under this framing, behavioral cloning (BC)~\citep{pomerleau1989alvinn} proposes to learn $\pi$ by minimizing
\begin{equation}
    \mathcal{L}_\text{BC}(\pi) \defeq \E_{(s,a)\sim\Demos}[-\log\pi(a|s)].
    \label{eq:bc}
\end{equation}
Equation \ref{eq:bc} can be viewed as the classification loss (discrete actions) or regression loss (continuous actions) of state to action mappings, connecting BC to supervised learning in vision and language.

\subsubsection{Reinforcement Learning} \label{sec:background-rl}
Standard reinforcement learning~\citep{sutton2018reinforcement} aims to maximize the expected returns of a policy through trial-and-error interaction with the environment:
\begin{equation}
\jrl(\pi) \defeq \E\left[\textstyle{\displaystyle}\left.\sum_{t=0}^H\gamma^t r_t\right|\pi,\mdp\right].\label{eq:maxret}
\end{equation}

\paragraph{Policy-based methods.} One conceptually straightforward way to optimize Equation \ref{eq:maxret} is through policy gradient, which estimates the gradient of Equation \ref{eq:maxret} with respect to the policy $\pi$, and maximizes $\jrl(\pi)$ directly via gradient ascent. The most commonly used gradient estimator has the form
\begin{equation}
    \nabla_\theta\jrl(\pi_\theta) = \E_{\tau\sim p_{\pi_\theta}(\tau)}\left[\textstyle{\displaystyle}\sum_{t=0}^H \gamma^t\nabla_\theta\log\pi_\theta(a_t|s_t)\hat{A}(s_t, a_t)\right],\label{eq:pg}
\end{equation}
where $\hat{A}$ is some advantage function that can be separately estimated via Monte-Carlo returns from $p_\pi(\tau)$~\citep{williams1992simple}. The biggest drawback of policy gradient is sample inefficiency: since policy gradients are estimated from rollouts, the variance of the gradient estimate is often extreme. To mitigate high variance, various works such as PPO~\citep{schulman2017proximal} have proposed to improve policy updates through the use of appropriate geometry~\citep{kakade2001natural,peters2010relative,schulman2015trust} or through training a separate critic network to estimate $\hat{A}$ to futher reduce variance at the cost of introducing bias~\citep{sutton1999policy,silver2014deterministic,schulman2015high}.

\paragraph{Value-based methods.} Another family of reinforcement learning methods for optimizing Equation \ref{eq:maxret}, such as Q-learning~\citep{watkins1992q}, involves learning the optimal value function $Q^*(s_t, a_t)$ by satisfying a set of Bellman \emph{optimality} constraints:
\begin{equation}
    Q^*(s_t, a_t) = r_t + \gamma \E_{s_{t+1}\sim\Trans(s_{t+1}|s_t,a_t)}\left[\textstyle{\displaystyle}\max_{a_{t+1}} Q^*(s_{t+1}, a_{t+1})\right],\label{eq:bellman-opt}
\end{equation}
after which an optimal policy can be extracted via $\pitarget(\cdot|s_t) = \arg_a \max Q^*(s_t, a)$. Value-based methods are typically more sample efficient than policy-based methods~\citep{gu2016q}, but tend to be unstable under function approximation~\citep{sutton2018reinforcement}. At the intersection of policy and value based methods, Actor-Critic methods~\citep{sutton1999policy} first learn $Q^\pi(s_t, a_t)$ by satisfying the set of Bellman \emph{expectation} constraints:
\begin{equation}
    Q^\pi(s_t, a_t) = r_t + \gamma \E_{s_{t+1}\sim\Trans(s_{t+1}|s_t,a_t),a_{t+1}\sim\pi(s_{t+1})}\left[\textstyle{\displaystyle} Q^\pi(s_{t+1}, a_{t+1})\right],\label{eq:bellman-exp}
\end{equation}
 then plug $\hat{A}(s_t, a_t) = Q^\pi(s_t, a_t)$ into the policy gradient objective, Equation \ref{eq:pg}, to update the policy. The intuition that the resulting policy learning will be both stable and sample efficient.

\paragraph{Off-policy and offline RL.} To further improve the sample efficiency of  on-policy methods, a set of off-policy approaches have been proposed for both policy and value based RL~\citep{lillicrap2015continuous,mnih2016asynchronous,nachumetal17}, where data from sources other than the current policy can be utilized for learning in conjunction with environment interaction. Offline RL~\citep{levine2020offline} further considers the setting where an agent only has access to a fixed dataset of previous interactions $\Doff$, and no further environment access to $\Trans$ or $\Reward$ is available. To ensure the learned policy avoids out-of-distribution states and actions, offline RL methods often impose regularization via a divergence between the learned policy and the offline dataset~\citep{wu2019behavior} or on the learned value function~\citep{kumar2020conservative}. 
More recently, some works have explored using additional online access as a finetuning step after offline RL to improve sample efficiency~\citep{nair2020accelerating,xie2021policy,ball2023efficient}. 

Using foundation models for decision making differs from traditional offline RL (with or without online finetuning) in that the latter focuses on learning RL algorithms from task-specific RL datasets $\Doff$ (i.e., datasets with task-specific states, actions, and rewards), whereas the former focuses on self-supervised learning on diverse data (e.g., data from vision and language domains) followed by task-specific adaptation.

\subsubsection{Planning, Search, and Optimal Control} 
Unlike the model-free RL algorithms outlined above, a broader set of approaches to sequential decision making (e.g., planning, search, optimization-based control, model-based RL) leverage explicit models of the environment. When the true environment dynamics are known (e.g., the rules of a Chess game) and simulation is cheap, planning and search algorithms, such as MCTS~\citep{kocsis2006improved} that leverage an accurate simulator, can be highly effective~\citep{silver2016mastering}. When the environment can be characterized by precise dynamics, such as the constrained movements of a robot arm, approaches in optimal control---such as 
trajectory optimization~\citep{von1992direct}, shooting~\cite{bock1984multiple}, collocation~\citep{von1993numerical}, and model predictive control~\citep{camacho2013model}---have long been studied prior to the recent advances in deep learning. In deterministic scenarios, given an environment governed by a known dynamics function $s_{t+1} = f(s_t, a_t)$, optimizing a sequence of actions $a_{0:T}$ to execute in the environment corresponds to  
\begin{equation}
    a_{0:T} = \arg \max_{a_{0:T}} J(s_{0}, a_{0:T}) = \arg \max_{a_{0:T}} \sum_{t=0}^T r(s_{t}, a_{t})\,\,\, \text{subject to}\,\, s_{t+1} = f(s_t, a_t). \label{eq:traj-opt}
\end{equation}
Model-based RL~\citep{doya2002multiple} considers the setting where the environment dynamics are unknown and have to be estimated from samples, after which techniques from search, planning, and optimal control~\citep{doya2002multiple,deisenroth2011pilco,tassa2012synthesis,nagabandi2018neural,kaiser2019model}
can be effectively applied given the learned dynamics model.

\subsection{Example Scenarios}
\begin{figure}
    \centering
    \includegraphics[width=0.9\textwidth]{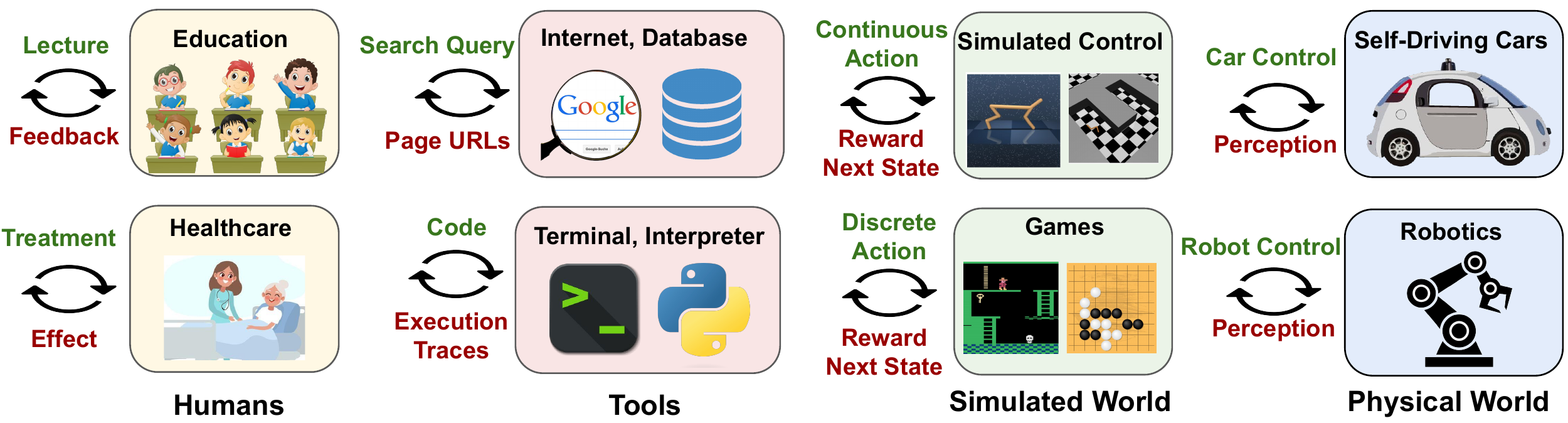}
    \caption{Example scenarios of adapting foundation models to perform decision making tasks such as interacting with humans, tools, and the simulated and physical world. Actions generated by foundation models and feedback provided by the external entities often reoccur repeatedly in a loop.}
    \label{fig:example}
\end{figure}
Before diving into the details of foundation models for decision making, we first discuss a few example scenarios where joint consideration of foundation models and decision making can be highly beneficial. Figure~\ref{fig:example} illustrates additional examples where foundation models can interact with external entities (e.g., humans, tools, and simulated and physical worlds).

\paragraph{Learning dialogue agents with human feedback.} There has been an increasing demand for large language models to produce likable, factual, and grounded responses to human inquires. With a moderate
amount of human feedback, via prompting or reward-based finetuning, langauge models have been able to perform increasingly more complex reasoning and dialogue tasks. Such feedback can be seen as the result of langauge model agents interacting with the external world (i.e., humans). Learning from interaction lies at the center of decision making, and reinforcement learning techniques such as policy gradient introduced in Section~\ref{sec:background-rl} have contributed significantly to the advances of dialogue systems \citep{ouyang2022training}.

\paragraph{The Internet as an environment.} 
While RL with human feedback has demonstrated tremendous empirical success in dialogue~\citep{thoppilan2022lamda,chatgpt2022}, humans are by no means the only external entity that can provide feedback to improve foundation models through repeated interaction. For instance, the Internet can be viewed as an unbounded environment where an ideal policy should be able to identify the best queries and navigation steps to retrieve  optimal answers in a minimal number of interactive steps. Since the Internet is both rich in information and cheap to interact with, it provides a compelling environment to explore decision making techniques. Foundation models are necessary for Internet-scale decision making, as interaction needs to be initiated in a reasonable way to ensure meaningful feedback is obtained for further learning.

\paragraph{Video generation as a universal policy.} 
A central difficulty in learning general-purpose robot agents is the incongruity between the state and action spaces of different environments. This implies that, for example, data collected by different robots cutting an apple or videos of a human cutting an apple cannot be easily combined to train a generalist robot policy, despite the fact that the notions of ``cutting'' and ``apple'' are common between these scenarios. With ever-larger text-to-video foundation models being trained on Internet-scale data~\citep{ho2022imagen,villegas2022phenaki}, it is now possible to recast the problem of policy learning as a text-conditioned video generation problem, where the generation process encompasses both environment modeling and planning. Such a policy-as-video formulation allows a unified interface (i.e., images) for learning and generalization from broad data sources, environments, and tasks.

\hypertarget{technical-generation}{\section{Foundation Models as Conditional Generative Models}}
\label{sec:technical-generation}

We now examine the first concrete use case of foundation models in decision making: probabilistic modeling of the trajectory distribution $p(\tau)$ from an interactive dataset $\tau\sim\Doff$. 
Depending on what part of $\tau$ is being modeled, foundation models can serve as conditional generative models of behaviors (i.e. actions) or the underlying world models (i.e., environment dynamics). Below, we first review different generative models and then discuss and explore how they can be used to represent behaviors and models of the environment.

\subsection{Generative Model Preliminaries}

Many foundation models can be characterized as modeling a (conditional) density $p(x)$ on a large dataset of images or texts $x\sim\Dset$. For example, $x$ could be an image, a sequence of images, or a sequence of text tokens. Different foundation models differ in their factorizations of $p(x)$. Below, we provide a brief overview of several generative models and their factorizations of $p(x)$.

\subsubsection{Latent Variable Models}
Latent variable models factorize the unknown data distribution of interest $p(x)$ into a latent variable distribution and a conditional distribution:
\begin{equation}
p(x) = \int p(z) p(x|z) dz,
\end{equation}
where the latent variable $z$ can be both discrete or continuous. For the special cases when $z$ is discrete and the sum is tractable, or $z$ is continuous and the integral is tractable, one can simply calculate $p(x)$ in closed form to support efficient maximum likelihood estimation on a given dataset. However, for the more general cases when the requisite sum or integral is intractable, techniques like VAEs~\citep{kingma2013auto} are applied to optimize the evidence lower-bound (ELBO) of $p(x)$ using a variational posterior $q(z|x)$:
\begin{equation}
    \mathcal{L}_\mathrm{VAE}(p,q) = \E_{x\sim\Dset,z\sim q(z|x)}\left[-\log p(x|z)\right] + \E_{x\sim\Dset}\left[\dkl\left(q(z|x)\|p(z)\right)\right]. \label{eq:vae}
\end{equation}
As an extension of VAE, VQ-VAE~\citep{van2017neural} uses a codebook to discretize the continuous latent representation to learn a more compact, discrete representation of the data.


\subsubsection{Autoregressive Sequence Models}
Autoregressive sequence models have been popularized by transformer-based language models~\citep{vaswani2017attention,brown2020language}. At their core, autoregressive models factorize any joint distribution over a sequence
$x = (x_1,...x_L)$ in an autoregressive manner:
\begin{equation}
p(x) = \Pi_{\ell=1}^L p(x_\ell|x_{<\ell}).
\end{equation}
Under this factorization, estimating the density $p(x)$ reduces to learning each conditional factor $p(x_\ell|x_{<\ell})$ which can be parametrized by a transformer.
\begin{equation}
    \mathcal{L}_\mathrm{LM}(p) = \E_{x\sim\Dset}\left[\sum_{\ell=1}^L -\log p(x_\ell|x_{<\ell})\right].\label{eq:lm}
\end{equation}


\subsubsection{Diffusion Models}
Diffusion models~\citep{sohl2015deep,ho2020denoising,kingma2021variational} are a class of latent variable models that factorize the data distribution $p(x)$ as a Markov chain of Gaussian transitions from a noise distribution of the same dimension:
\begin{equation}
    p(x) = \int p(x_K)\Pi_{k=1}^K p(x_{k-1}|x_k)dx_{1:K}, \label{eq:diffusion}
\end{equation}
where $p(x_K)=\mathcal{N}(\mathbf{0},\mathbf{I})$ and $p(x_{k-1}|x_k) := \mathcal{N}(\mu(x_k, k), \sigma(x_k, k))$. The forward diffusion process corrupts $x$ by iteratively adding Gaussian noise with a fixed variance schedule. The reverse process then achieves data generation by approximating the noise that corrupted $x$ during the forward process.


\subsubsection{Energy-Based Models}
Energy-based models~\citep{lecun2006tutorial, du2019implicit} are a class of models that represent data distributions $p(x)$ by an unnormalized distribution parameterized by a learned energy function:
\begin{equation}
    p(x) = \frac{e^{-E(x)}}{Z},
\end{equation}
where $E$ is the energy function and $Z = \int e^{-E(x)}dx$ is the partition function. To sample from the underlying distribution $p(x)$, one typically runs an MCMC procedure such as Langevin dynamics to sample from the underlying distribution.

\subsection{Generative Models of Behavior}
\begin{figure}
    \centering
    \includegraphics[width=\textwidth]{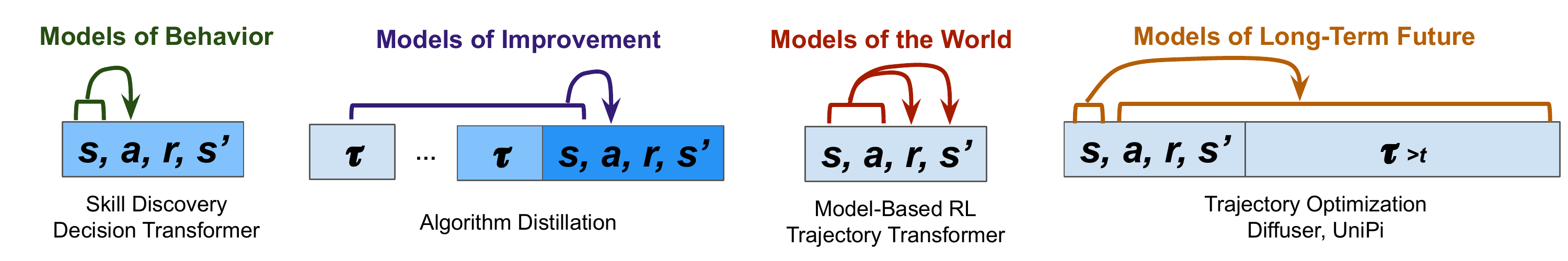}
    \caption{Illustrations of how conditional generative models can model behaviors, improvements, environments, and long-term futures given a trajectory $\tau\sim\Doff$. Dark blue indicates transitions with higher rewards. Models of behavior (Decision Transformers~\citep{lee2022multi}) and self-improvement (Algorithm Distillation~\citep{laskin2022context}) require near-expert data. Models of the world (Trajectory Transformer~\citep{janner2021offline}) and long-term future (UniPi~\citep{du2023learning}) generally require data with good coverage. 
    }
    \label{fig:generation}
\end{figure}

The generative models introduced above have mostly been applied to text or image data $x\sim\Dset$. Decision making, on the other hand, is concerned with  task specific \emph{interactive} data $\tau\sim\Doff$ that distinguishes state, action, and reward labels. We will see how different generative models can be adopted to model agent behaviors (this subsection) and environment dynamics (next subsection), as illustrated in Figure~\ref{fig:generation}.
\subsubsection{Foundation Models as Behavioral Priors}
When the interactive data $\Doff$ contains diverse behaviors such as ``pick up objects'', ``move objects horizontally'', or ``place objects'', these behaviors can be composed to complete tasks that were not present in $\Doff$. Foundation models can be used to model such ``behavioral priors''
(also known as ``skills'' or ``options'').  In this approach, pretraining generally involves maximum likelihood estimation of actions conditioned on some trajectory level information. Different tractable approximations can be leveraged to optimize this underlying training objective. For instance, the VAE objective from Equation \ref{eq:vae} can be directly instantiated, where the encoder $q$ takes a trajectory $\tau$ or some future goal as input and the decoder $\pi$ produces the sequence of actions as outputs~\citep{ajay2020opal,lynch2020learning}:
\begin{align}
    \label{eq:vae-bc}
    \mathcal{L}_\mathrm{VAE}(\pi, q) = \E_{\tau\sim\Doff,z\sim q(z|\tau)}\left[\sum_{t=0}^H -\log \pi (a_t|s_t,z) \right]+\E_{\tau\sim\Doff}\left[\dkl(q(z|\tau)\|p(z|s_0))\right].
\end{align}
The posterior distribution $q(z|\tau)$ can represent a diverse set of behavioral priors when $\tau$ is drawn from a wide set of related tasks. Since the posterior depends on future information, the prior $p(z|s_0)$ is usually constrained to only depend on the past so that behaviors can be correctly sampled at test time. 

Similarly, the autoregressive sequence modeling objective from Equation \ref{eq:lm} can also be instantiated to model behavioral priors~\citep{shafiullah2022behavior}, resulting in a policy that can depend on the history of interaction $\pi(a_t|s_t,\tau_{<t})$. Such dependence is less common in Markovian environments, but has shown empirical benefits~\citep{brohan2022rt}. When the dataset consists of expert data $\Doff^*$, one can learn transformer-based BC policies by optimizing the sequence modeling objective where an autoregressive transformer encodes the history $(\tau_{<t}, s_t)$ and decodes the next action $a_t$ as:
\begin{equation}
    \mathcal{L}_\text{LM}(\pi) = \E_{\tau\sim\Demos}[\sum_{t=0}^H-\log\pi(a_t|\tau_{<t},s_t)].
    \label{eq:bc-lm}
\end{equation}

An additional conditioning variable $z$ that captures trajectory-level information such as the goal or return $z(\tau) = \Return(\tau)$ has been introduced in goal or return conditioned supervised learning~\citep{schmidhuber2019reinforcement,kumar2019reward,brandfonbrener2022does,paster2022you,yang2022dichotomy}:
\begin{equation}
    \label{eq:lm-bc}
    \mathcal{L}_\mathrm{LM}(\pi) = \E_{\tau\sim\Doff}\left[\sum_{t=0}^H -\log \pi (a_t|\tau_{<t},s_t,z(\tau)) \right].
\end{equation}
When behavior generation is conditioned on high returns, 
intuitively, desirable behavior is encouraged~\citep{chen2021decision}. 

One can also utilize a diffusion model to model the conditional distribution of behaviors \citep{ajay2022conditional} by maximizing the likelihood in Equation \ref{eq:diffusion}:
\begin{equation}
    \label{eq:lm-bc}
    \mathcal{L}_\mathrm{Diffusion}(\pi) = \E_{\tau\sim\Doff, k\sim K}\left[\sum_{t=0}^H -\log \pi (a_t^{k-1}|a_t^k, s_t,z(\tau)) \right].
\end{equation}
To extract desirable behavior from a diffusion model when conditioned on high reward, one can sample trajectories with high likelihood 
by using reward as classifier-free guidance~\citep{ho2022classifier}.

Other conditional generative models that use normalizing flows~\citep{singh2020parrot}, generative adversarial networks~\citep{ho2016generative}, and energy-based models~\citep{florence2022implicit} have also been proposed for modeling behavioral priors from $\Doff$.

\subsubsection{Generalist Agents Trained on Massive Behavior Datasets}
A key advantage to generative modeling of behaviors lies in scaling up; despite different tasks possessing different observations and rewards, there are often meaningful behaviors shared across tasks
(e.g., ``moving left'' has similar meaning in navigation, game playing, and robot manipulation tasks). Inspired by the scaling success of transformers, generalist agents modeling sequences of diverse behaviors have been developed for simulated tasks~\citep{shafiullah2022behavior}, over 40 Atari games~\citep{lee2022multi}, over 700 real-world robot tasks~\citep{brohan2022rt}, and over 600 distinct tasks
with varying modalities, observations and action specifications~\citep{reed2022generalist}. 
This has led to generalist agents that are able to play video games, caption images, chat, perform robot tasks, 
significantly better than specialist agents trained on single tasks. Such works have also demonstrated the benefit of scaling model parameters and the number of training tasks.

While combining multiple task-specific datasets $\Doff$ into a large multi-task dataset as described above is one way to scale up behavior modeling, exploiting Internet-scale collections of text and video data $\Dset$ is another viable approach to scaling effectively. Internet-scale text and video data is abundant in quantity but typically has limited action annotations compared to $\Doff$.
Nevertheless, previous work has still incorporated such datasets. For instance, Gato~\citep{reed2022generalist} approaches this issue with universal tokenization, so that data with and without actions can be jointly trained using large sequence models. UniPi~\citep{du2023learning} directly learns to predict robotic videos and trains a separate inverse model to infer actions from generated videos. Applying inverse dynamics models to label large video data (e.g., from YouTube) is also applicable to other domains such as self-driving cars~\citep{zhang2022learning} and video game playing~\citep{baker2022video,venuto2022multi}.

\subsubsection{Large Scale Online Learning}
An alternative approach to assuming access to large-scale behavior datasets, 
online access to massive online game simulators
has enabled ``large-scale'' online RL models to be trained in games
such as DoTA~\citep{berner2019dota} and StarCraft~\citep{vinyals2019grandmaster}
using policy gradient or actor-critic algorithms.
Similarly, domain randomization~\citep{tobin2017domain} has been proposed to leverage online access to diverse generated environments to help bridge the sim-to-real gap in robotics. 
These large scale online training schemes, however, have not been able to leverage foundation models.
An important direction for future work is to explore how one can utilize and learn generative models similarly in massive online settings. 

\subsubsection{Generative Models of Exploration and Self-Improvement}
Generative models of behavior can also be extended to model meta-level processes, such as exploration and self-improvement, whenever the dataset itself $\Doff$ embodies exploratory and self-improving behavior (e.g., the replay buffer of a policy gradient agent trained from scratch)~\citep{laskin2022context}. 
That is, unlike other meta-RL methods, which usually train in online settings by maximizing multi-episodic value functions~\citep{wang2016learning,duan2016rl}, 
algorithm distillation imitates the action sequence of a multi-episodic improvement process from $\Doff$ by using a transformer-based sequence model inspired by the zero-shot ability of language models,
and adapts to downstream tasks purely in-context without updating any network parameters.

Similar to algorithm distillation, which prompts an agent with its prior learning experience, corrective re-prompting also treats long-horizon planning as an in-context learning problem, but uses corrective error information as prompts, essentially incorporating feedback from the environment as an auxiliary input to improve the executability of a derived plan~\citep{raman2022planning}.

\subsection{Generative Models of the World}\label{sec:gen-world}
In addition to learning models of behaviors, generative models can also learn models of the world---i.e., the transition dynamics $\Trans$ and the reward function $\Reward$---from the offline dataset $\Doff$. Conditional generation from a world model is analogous to model-based rollouts, which can be used to improve a policy.

\subsubsection{One-Step Prediction of Reward and Dynamics for Model-based Planning}
One can view learning models of $\Trans$ and $\Reward$ as a generative modeling problem given trajectories from an offline dataset $\tau\sim\Doff$. Since $\Doff$ also contains actions from a behavior policy $\pi$, then $\pi$, $\Trans$, and $\Reward$ can be jointly modeled with a single generative procedure. Specifically, the joint distribution of a trajectory $p(\tau)$ can be factored autoregressively into an environment component and a policy component,
\begin{equation}
p(\tau) = \Pi_{t=0}^H p(s_t, r_t, a_t|\tau_{<t}) = \Pi_{t=0}^H \Trans(s_t|\tau_{<t}) \cdot \pi(a_t|\tau_{<t}, s_t) \cdot \Reward(r_t|\tau_{<t}, s_t, a_t) ,\label{eq:lm-world}
\end{equation}
so that maximum likelihood estimation of $p(\tau)$ using $\Doff$ under this factorization naturally decomposes into learning the environment dynamics $\Trans, \Reward$ and the policy $\pi$ that produced the dataset $\Doff$.

Unlike language models where words exist in a common discrete space, here the states, actions and rewards in $\tau$ can all be expressed in different modalities, which poses challenges to sequentially modeling $\tau$. As a workaround, the Trajectory Transformer~\citep{janner2021offline} discretizes each dimension of states, actions, and rewards in a continuous control task before applying a GPT-style autoregressive model on the discretized tokens. Discretization is more challenging in image-based domains, where learning a latent representation of an image space and latent dynamics model is more common. Here one can introduce a per-step latent variable $z_t$ into the sequence modeling objective in Equation \ref{eq:lm-world}:
\begin{equation}
p(\tau) = \Pi_{t=0}^H\int_{z_t} \Trans_\textrm{enc}(z_t|\tau_{<t}) \cdot \Trans_\textrm{dec}(s_t|\tau_{<t}, z_t) \cdot \pi(a_t|\tau_{<t}, z_t)  \cdot \Reward(r_t|\tau_{<t}, z_t, a_t) dz_t,\label{eq:vae-world}
\end{equation}
where $\Trans_\textrm{enc}(z_t|\tau_{<t})$ encodes the history into the next step's latent state, $\Trans_\textrm{dec}(s_t|\tau_{<t}, z_t)$ decodes the next step's observation, and the policy $\pi$ and reward $\Reward$ can take latent state $z_t$ as input. Along this line, both \citet{hafner2020mastering} and \citet{chen2022transdreamer} apply a sequential VAE~\citep{zhu2020s3vae} to optimize the ELBO of Equation \ref{eq:vae-world}, and parametrize the latent dynamics model using an RNN or transformer based state space model respectively. Similarly, \citep{micheli2022transformers,ozair2021vector,seo2022reinforcement,seo2022masked} usesd VQ-VAE or masked autoencoders (MAE) to map image-based observations into discrete tokens before learning a transformer or latent state space dynamics model on the discretized observations.

The various ways a learned world model can be used to infer a high quality policy have been method and task specific. For example, heuristic decoding such as return guided beam search and MCTS have been applied to policy optimization \citep{janner2021offline,sun2022plate,ozair2021vector}. Separate actor and critic pairs have also been trained using rollouts from a latent world model (also referred to as ``imagination'') without requiring generating image-based observations~\citep{racaniere2017imagination,hafner2019dream}.
A world model, when trained to predict observations and actions in the original input space, can also be used to generate additional training data for model-free RL~\citep{sutton1990integrated,feinberg2018model,kaiser2019model,agarwal2020model} under the Dyna framework~\citep{sutton2018reinforcement} or to generate additional input context to a policy~\citep{du2019dynprior}.

\subsubsection{Planning with Generative Models of Long-term Future}
Instead of autoregressively factoring $\tau$ by time step as in Equation \ref{eq:lm-world}, one can also directly model the joint distribution of $\tau$ across all time steps at once using a diffusion model \citep{Du2019ModelBP,janner2022planning}:
\begin{equation}
    p(\tau) = p(s_0, a_0, r_0, \ldots, s_H, a_H, r_H) = \int p(\tau_K)\Pi_{k=1}^K p(\tau_{k-1}|\tau_k)d\tau_{1:K}.
\end{equation}
By learning a trajectory level generative model, planning can be more easily integrated with dynamics modelling by sampling from the composed distribution
\begin{equation}
    \tilde{p}(\tau) \propto p(\tau) z(\tau),\label{eq:world-diffusion-sample}
\end{equation}
where $z(\tau)$ specifies the trajectory-level properties that one wishes to control. For instance, \citet{janner2022planning} uses trajectory returns as $z(\tau)$ to guide a reverse diffusion process towards sampling high-return trajectories. \citet{ajay2022conditional} further demonstrate that $z(\tau)$ can represent different trajectory-level properties such as goals, skills, and dynamics constraints, where classifier-free guidance can be applied to conditionally sample trajectories that satisfy the desired properties. Going beyond low dimensional state action spaces, \citep{du2023learning} also show that diffusion models of long-term futures can also be applied to high-dimensional video data $\tau$, using $z(\tau)$ as text descriptions, effectively improving decision making with large-pretrained text-video foundation models.

In addition to the benefit of flexible conditioning (e.g., on returns, goals, constraints, skills, texts), sampling from the composed distribution in Equation~\ref{eq:world-diffusion-sample} holds the promise of accurate long horizon planning, since sampling an entire trajectory does not suffer from compounding error when rolling out single-step dynamics. Beyond diffusion models, EBMs can also be used to model the joint trajectory distributions $p(\tau)$, including conditioning on latent trajectory properties $z(\tau)$, which might provide a natural approach to satisfying multiple desirable properties, such as high return and safety \citep{du2020compositional,liu2022compositional}.





\hypertarget{technical-compression}{\section{Foundation Models as Representation Learners}}
\label{sec:technical-compression}

In this section, we discuss foundation models for decision making 
that leverage representation learning for
knowledge compression. 
On one hand, foundation models can extract representations from broad image and text data, $\Dset$, resulting in a plug-and-play style of knowledge transfer to vision and language based decision making tasks. On the other hand, foundation models can also be used to support task-specific representation learning via 
task-specific objectives and interactive data, $\Doff$.

\subsection{Plug-and-Play}
Off-the-shelf foundation models pretrained on Internet-scale text and image data can be used as preprocessors or initializers for various perceptual components of decision making agents. For instance, when an agent's perception is based on images, contrastive learning~\citep{chen2020simple} and masked autoencoding~\citep{he2022masked} 
can be directly applied to the agent's image observations, providing state representations that can be further finetuned by BC or RL objectives~\citep{sermanet2018time,kostrikov2020image,laskin2020curl,xiao2022masked}.
When agent actions can be characterized by natural language (e.g., ``move to the left then pick up the cup''), pretrained language models can be used to generate higher-level plans for longer-horizon tasks, with the hope that language based descriptions of actions generalize better than low-level motor controls~\citep{huang2022language,ahn2022can,wang2023describe,driess2023palme}.
When agent observations consist of both images and text descriptions, vision-language captioning models can further enrich agent observations with language descriptions~\citep{tam2022semantic,du2023guiding,driess2023palme}.
 Vision-language models such as CLIP and PaLI~\citep{chen2022pali} are further able to provide task feedback and reward information by aligning image and language modalities in the agent's observation and goal space~\citep{huang2022language,mahmoudieh2022zero,fan2022minedojo}. Even in the case where an agent's states, actions, and rewards do not consist of images or text, pretrained language models, perhaps surprisingly, have still been found useful as policy initializers for offline RL~\citep{reid2022can}, online RL~\citep{li2022pre}, and structured prediction tasks~\citep{lu2021pretrained}. 

Plug-and-play foundation models are 
generally more natural when the decision making task 
concerns real-world images or texts.
Plug-and-play is less applicable to decision making tasks 
when there are idiosyncratic, domain specific
state action spaces, which we will discuss in Section~\ref{sec:technical-compression-rl}. We will further discuss the challenges of bridging general image and text data with task-specific decision making data in Section~\ref{sec:challenge-data}.


\subsection{Vision and Language as Task Specifiers}
An important special case of plug-and-play foundation models is to use text commands or visual inputs as task specifiers to learn more robust, general, and multi-task policies
\citep{ahn2022can,huang2022language,brohan2022rt,liu2022instruction}. For instance, a text description of ``close the cabinet door'' or a goal image with the cabinet door closed can serve as policy input to augment the current robot state.
There are a few motivations behind this approach. First, using language and a goal image to specify a task provides richer information about the intended task rather than merely providing a scalar reward. Second, pretrained language models (equipped with prompting methods such as chain-of-thought) can decompose high-level tasks into lower-level instructions that are easier to execute~\citep{ahn2022can,huang2022language,jiang2022vima,team2021creating}. Furthermore, pretrained vision-language models can enable language-conditioned agents to generalize to new instructions, scenes, and objects in navigation and manipulation tasks~\citep{lynch2020language,hill2020human,hao2020towards,majumdar2020improving,nair2022learning,jang2022bc,ahn2022can,huang2022language,khandelwal2022simple,shridhar2022cliport,guhur2022instruction,shah2022lm}, which has been a key challenge in robotics prior to 
their introduction~\citep{zhu2018reinforcement}. 

Using vision and language task specifiers to prompt for desirable agent behaviors requires additional data such as text descriptions or goal images of a given task (see challenges in Section~\ref{sec:challenge-data}). Moreover, prompting for desirable outcomes from a large language model has significant potential but is also an open problem in itself~\citep{liu2023pre}, whose complexity is exacerbated in decision making scenarios with external entities and world dynamics (see Section~\ref{sec:challenge-dm}).

\subsection{Learning Representations for Sequential Decision Making} \label{sec:technical-compression-rl}

\begin{figure}[b]
    \centering
    \includegraphics[width=\textwidth]{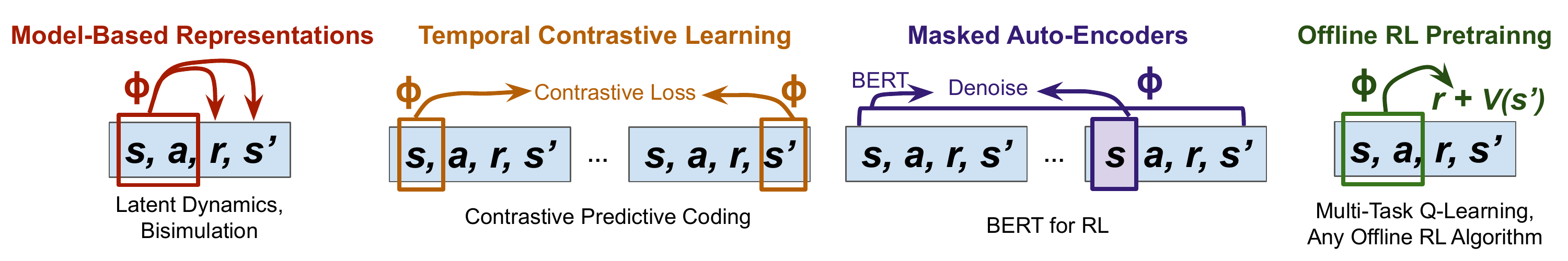}
    \caption{Illustrations of different representation learning objectives such as model-based representations~\citep{nachum2021provable}, temporal contrastive learning~\citep{oord2018representation}, masked autoencoders~\citep{devlin2018bert}, and offline RL~\citep{kumar2022offline}, on a trajectory $\tau\sim\Doff$ specifically devised for sequential decision making.
    }
    \vspace{-10pt}
    \label{fig:compression}
\end{figure}

Unlike vision-language foundation models that can learn from a broad data collection $\Dset$ but lack the notion of decision making, foundation model techniques and architectures (as opposed to the pretrained models themeselves) can be used to optimize objectives uniquely devised for sequential decision making on the basis of task-specific interactive data $\Doff$. Figure~\ref{fig:compression} visually illustrates these representation learning objectives.

\paragraph{Model-based representations.} Traditionally, representation learning for sequential decision making has been framed as learning a latent state or action space of an environment by ``clustering'' states and actions that yield similar transition dynamics~\citep{dearden1997abstraction,andre2002state,mannor2004dynamic,abel2018state,gelada2019deepmdp,agarwal2021contrastive}. Similar to how foundation models can serve as generative models of world dynamics by maximizing $p(\tau)$ in Equation \ref{eq:lm-world}, foundation models can also serve as representation learners of world dynamics under the following objective:
\begin{equation}
p(\tau_{s,r}) = \Pi_{t=0}^H p(s_{t+1}, r_t|\tau_{<t}, s_t, a_t) \\
= \Pi_{t=0}^H \Trans(s_{t+1}|\tau_{<t}, \phi(s_t), a_t) \cdot \Reward(r_t|\tau_{<t}, \phi(s_t), a_t)\label{eq:repr-world}.
\end{equation}
Using this factorization for maximum likelihood estimation of $p(\tau_{s,r})$ using $\Doff$ 
naturally leads to learning state representations $\phi(s)$ that ``cluster'' states with similar rewards and next state probabilities. One could also choose to maximize the likelihood of the next state \emph{representations} as opposed to the next raw state, i.e., $\Trans(\phi(s_{t+1})|\tau_{<t}, \phi(s_t), a_t)$ resulting in 
a latent dynamics model \citep{gelada2019deepmdp}. Alternative learning objectives for $\phi(s)$ can be derived depending on how $\Trans(s_{t+1}|\tau_{<t}, \phi(s_t), a_t)$ is defined. For instance, $\Trans$ may be defined as an energy-based model:
\begin{equation}
    \Trans(s_{t+1} | \tau_{<t}, \phi(s_t),a_t)\propto\exp\{\phi(s_{t+1})^\top f(\phi(s_t),a_t,\tau_{<t})\},\label{eq:repr-ebm}
\end{equation}
where $f$ is a trainable function that maps $\phi(s_t),a_t,\tau_{<t}$ to the same embedding space as $\phi$
. While Equation \ref{eq:repr-world} learns state representations by modeling the forward dynamics, one can also learn state representations based on an \emph{inverse} dynamics model~\citep{pathak2017curiosity,shelhamer2016loss} by predicting $a_t$ from $\tau_{<t}, s_t, s_{t+1}$, thereby maximizing
\begin{equation}
p(\tau_{a}) = \Pi_{t=0}^H p(a_t|\tau_{<t}, \phi(s_t), \phi(s_{t+1})).\label{eq:repr-inv}
\end{equation}
In addition to forward and inverse dynamics based representations, it is also possible to learn state representations derived from predicted value functions~\citep{oh2017value}, curiosity metrics~\citep{du2021curious}, or other MDP-based similarity metrics such as bisimulation properties deduced from Bellman backups~\citep{ferns2004metrics,castro2010using,zhang2020learning}. The above representation learning objectives have mostly been considered under the Markovian setting, hence the dependence on $\tau_{<t}$ is often dropped.  Though the Markovian assumption makes large sequence models seem less relevant, these representation learning objectives benefit from sequence modeling architectures in image-based domains that are generally non-Markovian.

\paragraph{Temporal contrastive learning.} The model-based representation objectives above require strictly interleaved state-action-reward tuples in the training data $\Doff$, which can preclude more flexible representation learning techniques that consider broader data sources, $\Dset$, such as YouTube videos (which can be thought of as state-only trajectories $\tau_s$). Temporal contrastive learning such as CPC~\citep{oord2018representation}, on the other hand, can model more flexible sequence-level representations, and has been applied to playing games by watching YouTube videos~\citep{aytar2018playing}. Specifically, in temporal contrastive learning, observations that are closer temporally (e.g., observations that belong to the same trajectory) are encouraged to have similar representations. Given a sub-trajectory $\tau_{t:t+h}$, one can learn $\phi(s)$ by minimizing a contrastive loss between $\phi(s_t)$ and $\phi(s_{t+i})$:
\begin{equation}
-\phi(s_{t+i})^\top W_i \phi(s_t) + \log \E_\rho [\exp\{\phi(\tilde{s})^\top W_i \phi(s_t)\}].\label{eq:cpc}
\end{equation}
where $i=1,\dots,h$, $W_i$ is a learnable weight matrix, and $\rho$ is some non-trainable prior distribution. Note that the temporal contrastive learning in Equation \ref{eq:cpc} bears resemblance to learning an energy-based dynamics model in Equation \ref{eq:repr-ebm}, as established in prior work~\citep{nachum2021provable,nguyen2021temporal}.

\paragraph{Masked autoencoders.} When a trajectory $\tau = (s_0, a_0, r_0,..., s_H, a_H, r_H)$ from $\Doff$ is treated as a flattened sequence, BERT-style denoising autoencoding objectives can be applied to the sequence to learn representations of states, actions, rewards, and dynamics through specific choices of masking patterns~\citep{yang2021representation,liu2022masked,carroll2022unimask,seo2022masked}. These methods learn representations $\phi(s)$ by first randomly masking a subset of tokens in $\tau$ to obtain $\hat\tau$, then pass the masked sequence $\hat\tau$ to a transformer, and finally reconstruct the masked portions of the original input $\bar\tau$ from the transformer output $F(\hat\tau)$. The training objective, for instance, can be characterized as maximizing
\begin{equation}
    p(\bar\tau | \hat\tau) = \Pi_{t=0}^{H} m_t p(\tau_t|\hat\tau) = \Pi_{t=0}^{H} m_t \frac{\exp\{F(\hat\tau)_t^T \phi(s_t)\}}{\sum_{s}\exp\{F(\hat\tau)_t^T\phi(s)\}},
\end{equation}
where for each masked input state $s_t$, a contrastive loss between its representation $\phi(s_t)$ and the transformer output at its sequential position $F(\hat\tau)_t$ is applied. Unlike model-based representation learning approaches that explicitly model state transition probabilities, masked autoencoders can learn representations from a broader dataset that potentially has missing actions and rewards, while still being able to incorporate dynamics-based information in the learned representations.

\paragraph{Offline RL pretraining.} When the downstream decision making tasks are to be trained with RL objectives, it might seem natural to apply similar RL objectives during pretraining when acquiring value-based representations~\citep{mazoure2022contrastive,ball2023efficient}. At a high level, value-based pretraining encompasses any offline RL algorithms that have been pretrained on logged experience from one or more tasks relevant to the downstream interactive task of interest. Value-based pretraining has exhibited scaling capability in multi-task settings where state action spaces are similar (e.g., all of Atari games~\citep{kumar2022offline}). 

\subsubsection{Post Representation Learning: BC and RL Finetuning}
Unlike generative foundation models that can directly produce action or next state samples, as in Section~\ref{sec:technical-generation}, foundation models as representation learners are only directed to extract representations of states, actions, and dynamics; hence they require additional finetuning or model-based policy optimization to achieve strong decision making performance. On the theoretical side, various works have focused on developing representation learning objectives that ensure downstream BC or policy/value-based RL finetuning using the pretrained representations are provably efficient~\citep{jin2020provably,nachum2021provable,zhang2022making,pacchiano2022joint,ren2022latent}. These analyses are generally based on properties of linear MDPs. For instance, one such assumption states that the state-action value function $Q^\pi(s, a)$ can be represented as a linear combination of features $\phi(s,a)$ under the linear MDP factorization $\Trans(s'|s,a) = \langle\phi(s,a),\theta(s')\rangle$ and $\Reward(s,a) = \langle\phi(s,a),\theta_r\rangle$, which ensures that standard policy and value based RL training can take place in the more compact representation space $\phi(s,a)$ as opposed to the original state-action space. Beyond providing compact state action spaces for policy and value-based model-free RL methods, pretrained representations can also simplify model learning and policy rollouts of model-based policy optimization~\citep{silver2014deterministic,oh2017value,hafner2019dream} as described in Section~\ref{sec:gen-world}. 

While representation learning objectives specifically devised for sequential decision making have theoretical benefits, it is less clear how these objectives can effectively incorporate broader and multi-task data when the underlying dynamics differ from that of the target task of interest. The recurring challenge of bridging learning from broad data $\Dset$ and task-specific data $\Doff$ will be further discussed in Section~\ref{sec:challenge-data}.

\hypertarget{technical-lm}{\section{Large Language Models as Agents and Environments}}
\label{sec:technical-lm}
We have seen that foundation models can characterize different components of a decision making process ($\mdp$), such as agent behaviors ($\Aset$), world dynamics ($\Trans$), task specifiers ($\Reward$), and state ($\Sset$) and action representations. In this section, we further consider a special case where pretrained large language models can serve as agents or environments. Treating language models as agents, on one hand, enables learning from environment feedback produced by humans, tools, or the real world, and on the other hand enables new applications such as information retrieval and web navigation to be considered under a sequential decision making framework. Language models can also be thought of as computational environments that take text as input and produce text as output, effectively supporting interactions with external prompts.

\subsection{Interacting with Humans}
\paragraph{Dialogue as an MDP.} A piece of dialogue can be viewed as in alternating nteraction between a dialogue agent $\pi$ and a human environment $\mdp=\mathcal{E}$, where a conversation $\tau_{<t}=\{e_0,a_1,e_1,...,a_t\}$ consists of sentences $a_i$ and $e_i$ produced by $\pi$ and $\mathcal{E}$ respectively. On the $t$-th turn, a state $s_t \in \Sset$ captures the conversation history $s_t=\{\tau_{<t}, e_t\}$, an action $a_t \in \Aset$ is an agent's response given this context, a next state $s_{t+1} \in \Sset$ concatenates $s_{t}$ with $a_t$ and $e_{t+1}$, and a reward $r_t=\Reward(s_t,a_t)$ is produced. An agent $\pi$ aims to maximize $\E_{e_0 \sim \init, \pi,\Trans} [\sum_{t=0}^{H}\gamma^t \Reward(s_t,a_t)]$.

\paragraph{Optimizing dialogue agents.} The application of large language models to dialogue generation is a natural one, as both the broad pretraining data $\Dset$ and the task-specific dialogue data $\Doff$ are of the same text modality, which allows for task-specific finetuning using the same self-supervised loss as pretraining~\citep{adiwardana2020towards,roller-etal-2021-recipes,nakano2021webgpt,thoppilan2022lamda}. Such an approach has achieved impressive performance as assessed by humans, under metrics including safety, sensibleness, interestingness, truthfulness, and helpfulness \citep{thoppilan2022lamda,bai2022training}. Although human feedback was initially used to evaluate dialogue systems~\citep{jiang2021towards}, it was soon incorporated as a reward signal for optimizing dialogue agents under the \textit{reinforcement learning with human feedback} (RLHF) framework \citep[][\textit{inter alia}]{ouyang2022training,chatgpt2022,bai2022training}. In practice, RLHF involves several stages: first, a pretrained language model is finetuned on dialogue data to provide an initial policy $\pi$; second, output from this model is ranked by human raters, which is then used to train a preference (reward) model $\Reward$; finally, the language model is finetuned using policy gradient in Equation \ref{eq:pg} to maximize the reward given by the preference model. Other RL objectives such as Q-learning (Equation \ref{eq:bellman-opt}) and actor-critic (Equation \ref{eq:bellman-exp}) have also been used to enable dialogue agent to perform specific tasks, such as booking flights and selling items on Craigslist \citep{jaques2017sequence,verma2022chai,snell2022context,jang2022gptcritic,snell2022offline}.

\paragraph{Limitations of dialogue agents.} While using human feedback is a natural way to turn broad data $\Dset$ into task-specific data $\Doff$, solely relying on human feedback to finetune a language model agent has a number of limitations. For instance, language models have been criticized for failing to access up-to-date information~\citep{komeili2021internet}, hallucinating facts~\citep{maynez2020faithfulness,ji2022survey}, and struggling to perform complex reasoning and mathematical calculations~\citep{patel2021nlp}. Such failure modes are unsuprising if these desired properties were never a part of the feedback the language model received. While one approach to mitigate such failure modes is to collect human feedback on each of the desired properties, leveraging tools and external entities that can automatically provide feedback is likely to be a more scalable and reliable approach.

\subsection{Interacting with Tools}
Language model agents that generate API calls (to invoke external tools and receive responses as feedback to support subsequent interaction) can be formulated as a sequential decision making problem analogous to the dialogue formulation in the previous section. 
Several tools such as search engines~\citep{komeili2021internet,thoppilan2022lamda, lazaridou2022internet,shuster2022blenderbot,yao2022react}, calculators~\citep{cobbe2021training,thoppilan2022lamda}, translators~\citep{thoppilan2022lamda}, MuJoCo simulators~\citep{liu2022mind}, scratch pads~\citep{nye2021show}, computer memory~\citep{schuurmans2023memory}, and program interpreters~\citep{gao2022pal} have been used to augment language models in a supervised finetuning or prompting setting, where response from tools are used as additional inputs to the language model. 

\paragraph{Limitations of tool use agents.} Unlike dialogue systems, where the agent and environment take turns, tool-using agents need to additionally decide when to call external tools, which tools to use, and how to use these tools (e.g., reformulating query if results are not helpful), all of which pose additional challenges. Consequently, the supervised finetuning of tool-use agents requires significant human supervision through API call annotations. While prompting-based tool-use requires fewer examples, the specific prompts typically need to be hand-crafted for each tool~\citep{schick2023toolformer}. Moreover, language models are known to be sensitive to the prompt formats in both the zero and few-shot settings \citep{jiang-etal-2020-know,schick-schutze-2021-exploiting}. As a result, the communication between language models and external tools typically needs to be cleaned-up by a rule-based parser, which further complicates the prompting setup. Recently, \citet{parisi2022talm} and \citet{schick2023toolformer} have made progress on self-supervised learning of tool use with language models, training the language model to only an external tool if this leads to an improved response over the outcome predicted by language model alone. Nevertheless, none of the existing work considers tool use in an interactive setting where an agent can \emph{iterate} on its behavior according to tool feedback to improve its tool-use ability.

\paragraph{Tools as interactive environments.} 
It is challenging to scale
supervised finetuning and prompting 
to a large number of tools with different uses and tools that return large amounts of feedback (e.g., hundreds of search results). One sensible way of tackling this challenge is to treat tools like web browsers as interactive environments, from which experience can be sampled by executing search queries \citep{nakano2021webgpt,gur2022understanding}, and optimizing such queries via RL techniques such as policy gradient. Treating tools as interactive environments enables methods that require massive and efficient online simulator access (e.g., Monte Carlo Tree Search for AlphaGo) to be applied to a broader set of real-world problems, such as web navigation and information retrieval. Additionally, situating language models in true knowledge obtained from the environment better grounds the model, avoiding the the Dichotomy of Control problem (e.g., sequence models generating next states without respecting environment transitions)~\citep{yang2022dichotomy}.

\subsection{Language Models as Environments}

\paragraph{Prompting as an MDP.} Iterative prompting can be characterized as an MDP that captures the interaction between a prompt provider $\pi$ and a language model environment $\mathcal{E}$, where a prompt history $\tau_{<t}=\{e_0,a_1,e_1,...,a_t\}$ consists of prompts $a_i$ and language model outputs $e_i$ produced by $\pi$ and $\mathcal{E}$ respectively.  Here, $e_0$ is the initial context to the language model. In the $t$-th turn, a state $s_t \in \Sset$ captures the prompting history and the $t$-th language model responses $s_t=\{\tau_{<t}, e_t\}$, an action $a_t \in \Aset$ is given by the prompt provider, a next state $s_{t+1} \in \Sset$ is produced by concatenating $s_{t}$ with $a_t$ and the next response of the language model $e_{t+1}$, and a reward $r_t=\Reward(s_t,a_t)$ is emitted. An agent $\pi$ aims to maximize $\E_{e_0 \sim \init, \pi,\Trans} [\sum_{t=0}^{H}\gamma^t \Reward(s_t,a_t)]$. In language model reasoning, for instance, $\Reward(s_t, a_t)=1$ if the language model's output successfully reaches a goal answer $s_t$ (i.e., correct reasoning), and $\Reward(s_t, a_t)=0$ otherwise.

Under this formulation, various schemes for language model prompting can be characterized by high-level actions that map input strings to desired output strings using the language model. For instance, such high-level actions include \texttt{DECOMPOSE}~\citep{press2022measuring}, \texttt{RANK}~\citep{kumar2021reordering}, \texttt{DENOISE}~\citep{shi2023large}, and \texttt{PARAPHRASE}~\citep{jiang2021can}. These high-level actions can also be recursively composed to achieve more sophisticated iterative prompting schemes~\citep{zhou2022least}. Other prompting schemes such as \texttt{SUMMARIZE}, \texttt{PRUNE}, \texttt{SEARCH} can be considered for handling challenges such as overcoming long context lengths. Given that language models with auxiliary memory have been shown to emulate universal Turing machines~\citep{schuurmans2023memory}, language models could ultimately serve as ``computers'' that also operate on human language with prompting as a flexible new form of programming language.



\hypertarget{challenge}{\section{Open Problems, Challenges, and Opportunities}}
\label{sec:challenge}

\subsection{How to Leverage or Collect Datasets}\label{sec:challenge-data}
One key challenge in applying foundation models to decision making lies in the dataset gap: the broad datasets from vision and language $\Dset$ and the task specific interactive datasets $\Doff$ can be of distinct modalities and structures. For instance, when $\Dset$ consists of videos, it generally does not contain explicit action labels indicating the cause-effect relationship between different frames, nor does it contain explicit reward labels indicating which videos are better than others, whereas actions and rewards are key components of $\Doff$. Despite this gap, broad video and text data can be made more task specific through post-processing ($\Dset\rightarrow\Doff$), leveraging hindsight relabeling of actions and rewards (e.g., using human feedback). Meanwhile, decision making datasets can be made more broad and general ($\Doff\rightarrow\Dset$) by combining a wide range of tasks-specific datasets (e.g., Gato). Below we provide a list of examples of $\Dset$ and $\Doff$ that can be used for research in foundation models for decision making, and propose additional approaches for bridging the gap between $\Dset$ and $\Doff$.

\paragraph{Existing vision and language datasets ($\Dset$).} Vision and language datasets can be useful for decision making if they contain multiple modalities (e.g., aligned image and text pairs), (implicit) actions, movements, instructions, and notions of tasks. For instance:
\begin{itemize}
    \item LAION-5B~\citep{schuhmann2022laion} contains 5.85 billion CLIP-filtered text-image pairs.
    \item Egocentric 4D Perception (EGO4D)~\citep{grauman2022ego4d} contains over 30k hours of time-aligned video in an inertial measurement unit (IMU) dataset of people's activities such as cooking, eating, and working at a computer in 4D (3D spatial and time).
    \item Something-Something V2 Dataset~\citep{goyal2017something} contains 220k short videos of people performing various tasks with everyday objects, such as putting on a hat and opening a bottle. These videos are annotated with action labels at the level of verb and noun phrases.
    \item HowTo100M~\citep{miech2019howto100m} contains over 100 million video clips and descriptive captions, covering topics such as cooking, home improvement, and beauty.
    \item BigBench~\citep{srivastava2022beyond} is a dataset consisting of NLP tasks such as question answering, summarization, and conversation modeling. It also contains text-based games such as text navigation, Sudoku, and Taboo.
\end{itemize}

\paragraph{Existing decision making datasets ($\Doff$).} Foundation models are currently relevant to decision making datasets that are larger-scale, multi-task, multi-modal, real-world based, and video or text based. For example:
\begin{itemize}
    \item BabyAI~\citep{chevalier2018babyai} contains data in text-based games that require an agent to navigate in a 2D gridworld virtual environment and perform a variety of tasks.
    \item VirtualHome~\citep{puig2018virtualhome} contains over 15k simulated images and videos of indoor scenes, along with detailed information of the scenes and objects such as object shape, size, and material properties.
    \item RoboNet~\citep{dasari2019robonet} contains over 100k video clips of 7 robots over 100 camera viewpoints performing a variety of tasks in different environments.
    \item RL Unplugged~\citep{gulcehre2020rl} is an offline RL dataset consisting of simulated locomotion, manipulation, and Atari games.    
    \item Bridge Data~\citep{ebert2021bridge} contains 7,200 text-video demonstrations of a 6-dof WidowX250s robot arm performing 71 tasks across 10 kitchen-themed environments.
    \item MineDojo~\citep{fan2022minedojo} contains 640k text-video pairs (16s in length), 7k Wiki pages, and 340k Reddit posts on Minecraft.    
    \item RT-1~\citep{brohan2022rt} Robotics Transformer for Real-World Control at Scale (to be released).
    \item CACTI~\citep{mandi2022cacti}: A Framework for Scalable Multi-Task Multi-Scene Visual Imitation Learning (to be released).
    \item VIMA~\citep{jiang2022vima} contains 650K successful trajectories of 17 simulated robotic manipulation tasks with interleaved language and image/video frames.   
\end{itemize}

\paragraph{Bridging $\Dset$ and $\Doff$.} To enable better datasets tailored for decision making, one can either increase the scale of $\Doff$ by large-scale logging and merging task-specific sets of interactive data, or by relabeling $\Dset$ with action and reward information. One could also consider augmenting $\Doff$ with meta data, such as informational and instructional texts and videos.
\begin{itemize}
    \item Large-scale logging of interactions. Since many automatable tasks are currently conducted by humans (driving, navigating the web, writing code), it is possible to collect large amounts of data for sequential decision making by logging human behaviors. Similar to logged human conversations that are used to train dialogue agents, one can log ``actions'' such as keystrokes and mouse movements for training web navigating agents.
    \item Hindsight relabelling of existing data. Since many videos are already available on YouTube, it is possible to relabel the videos in hindsight with task descriptions and action information similar to \citet{behbahani2019learning,shaw2022videodex}.
    \item Incorporating descriptions, instructions, and other task information. Since training a DQN Atari agent from scratch requires 7 GPU days, it is natural to consider whether information about an Atari game on the Internet (e.g., the Gameplay section of a game's Wikipedia page) could improve an agent's learning speed and sample efficiency.
\end{itemize}

\subsection{How to Structure Environments and Tasks}\label{sec:challenge-structure}
Foundation models in vision and language can often solve a diverse set of tasks and generalize to new tasks in a few-shot or zero-shot manner~\citep{radford2021learning,alayrac2022flamingo,brown2020language,chowdhery2022palm,hoffmann2022training}. Unlike vision and language where images or texts can serve as a universal task interface, decision making faces environment diversity where different environments operate under distinct state action spaces (e.g., the joint space and continuous controls in MuJoCo are fundamentally different from the image space and discrete actions in Atari), thereby preventing knowledge sharing and generalization. Below are some recent approaches to structuring environments and tasks so that foundation model architectures (e.g., Transformers) and large pretrained models (e.g., video diffusion) can be applied to decision making.
\begin{itemize}
    \item \textbf{Universal encoding.} Similar to \citet{reed2022generalist} and \citet{janner2021offline}, all states, actions, and rewards across different environments and tasks can be encoded into universal tokens in a sequence modeling framework. However, such tokenization might not be able to preserve the rich knowledge and generalization abilities of pretrained vision and language models.
    \item \textbf{Text as environment.} Alternatively, one can convert environments with different state action spaces into text descriptions and use text as a universal interface to learn generalizable policies. For instance, when an observation is an image, one may use a caption model to convert the observation to text, or directly use ASCII characters to represent the observation as text. Text-as-environment and LM-as-policy have been evaluated on a variety of simple interactive games such as Spelling Bee, Sudoku, Chess, and Taboo \citep{srivastava2022beyond}, though there is still a substantial gap between large language models and state-of-the-art task-specific game-solving systems (e.g., AlphaGo) in these tasks. Text as environment also seems unnatural in visual perception based applications such as self-driving. Instead of using text as states and actions, one can also use text descriptions to specify tasks (rewards)~\citep{ahn2022can,huang2022language,brohan2022rt,du2023learning}, avoiding the difficulties around reward shaping. Using text as a task specifier requires additional data to be collected, and still faces the challenge of incongruent state action spaces across tasks.
    \item \textbf{Video as policy and world model.} Finally, one can use image frames as a universal interface to represent state action spaces, and use videos to represent policies \citep{du2023learning}. This allows policy learning to leverage web-scale pretrained text-to-video models. However, the mapping from videos to joint actions of individual agents still requires further training. This approach is further complicated by the computational difficulty of effective video generative modeling.

\end{itemize}

\subsection{Improving Foundation Models}\label{sec:challenge-fm}
\paragraph{Long-context and External Memory} 
Effective decision making often requires long context of the prior history of observations and actions.  In contrast, existing approaches typically rely on transformers that have a bounded context length. To emulate general-purpose computations and decision making, properly incorporating interactions with external memory is important. One approach is to leverage prompting of intermediate computations  ~\citep{schuurmans2023memory,giannou2023looped} to extend computational context, but this approach is difficult to implement in practice due to the sensitivity of language models on prompt selection and ways of parsing the output. Another interesting direction for future exploration is to incorporate retrieval of past observations to enable effective decision making~\citep{borgeaud2021improving}.

\paragraph{Combining multiple foundation models.} Different foundation models capture different data modalities, such as visual, textual, and cross-modal representations of data. To effectively execute decision making across different environments, it is desirable to jointly leverage information across different models.  One approach to compose models across different modalities  is to graft them~\citep{alayrac2022flamingo} on top of a single large language model. Alternatively, language can be used as a ubiquitous interface in which separate foundation models can communicate ~\citep{zeng2022socratic}. Different foundation models can further communicate through iterative optimization~\citep{li2022composing}. A limitation of existing works is that they either require finetuning~\citep{alayrac2022flamingo} or defined interfaces within which models can communicate \citep{zeng2022socratic,li2022composing}, which prevents novel combinations of foundation models from being easily composed at test-time in a free-form manner.

\paragraph{Grounding foundation models in the world.} Foundation models are typically trained on Internet-scale data without knowledge of the physical world. To effectively execute actions produced by foundation models in the real world, it is important to ground these models in both the underlying geometry and physics of the world. One existing approach uses intermediate outputs from simulators as context for action generation~\citep{liu2022mind}. Alternatively, foundation model outputs could be scored and optimized using feedback from simulators~\citep{li2022composing}. Existing works assume access to a simulator of the operating environment, which is not available in the physical world. Constructing systems that more accurately ground predictions in the physical world is therefore an interesting area for future research.

\subsection{Improving Decision Making}\label{sec:challenge-dm}

\paragraph{How to extract desirable behavior.} 
One key aspect of foundation models for decision making lies in effectively adapting task-agnostic models into task-specific agents. Various approaches can be seen as ways to ``control'' foundation models to produce desirable behaviors for specific tasks. For instance, large-pretrained language models can be specialized to output desired sentences through instruction finetuning~\citep{wei2021finetuned} or few-shot prompting~\citep{brown2020language}. For conditional generative modeling of behavior, language goals~\citep{du2023learning}, image goals~\citep{brohan2022rt}, returns~\citep{lee2022multi}, environment constraints~\citep{ajay2022conditional}, and expert demonstrations~\citep{reed2022generalist} have all been explored as s conditioning factor for finetuning or prompting schemes, so that the models can be ``controlled''.

Aside from goal or instruction conditioned finetuning or prompting, two types of ``iterative'' approaches have also been applied to elicit expert behavior. The first approach iterates through a set of chain-of-thought reasoning or computation steps~\citep{nye2021show,wei2022chain,yang2022chain}, with the hope that a sequence model supervised to emit similar chain-of-thought steps will achieve better generalization. The second approach iterates through a set of improvement steps from less to more desirable behaviors, with the hope that a sequence model supervised on the improvement sequence can continue to regress on the improvement trend~\citep{laskin2022context,liu2023languages}. Both of these approaches, together with goal conditioned supervised learning, can help extract desirable behavior without requiring explicit finetuning with RL objectives.


\paragraph{Offline to online.} 
While conditional generative modeling can elicit expert behavior as discussed above, directly finetuning foundation model agents using RL objectives such as policy gradient is another approach. One major challenge that has prevented wide real-world adoption of RL finetuning is the need for large online samples to ensure learning progress~\citep{li2019reinforcement}. Nevertheless, in game settings where massive online access is available (e.g., Go, Chess, Shogi, Dota, Atari), RL methods have surpassed human performance. Instead of avoiding online access altogether through offline RL or conditional generative modeling, language models as interactive agents enables massive online access to environments that are highly scalable and available (e.g., search engines, databases, compilers). Developing infrastructures that enable software tools as environments, remote procedure calls as interactions, and foundation models as policies can have a large impact on a wide range of real-world applications.


\hypertarget{conclusion}{\section{Discussion and Perspectives}}
\label{sec:conclusion}


Foundation models have achieved remarkable success in emulating human intelligence at earlier stages of development: seeing, hearing, speaking, reading, and writing. To transform these basic human abilities to world-class expertise, humans spend tens of thousands of hours practicing through trial and error~\citep{gladwell2008outliers}, interacting with the external world, making mistakes, and learning from them. Foundation models for decision making offers a path to transform general artificial intelligence capabilities in vision, language, and world knowledge into next-level expert capabilities.

As well as achieving more sophisticated intelligence, foundation models can also characterize different components of a decision making system, such as generative models of behavior and the world (Section~\ref{sec:technical-generation}), representations of world knowledge (Section~\ref{sec:technical-compression}), and interactive agents or environments through the usage of language (Section~\ref{sec:technical-lm}). Despite the initial successes, foundation models for decision making inevitably faces significant challenges, such as the gap in data modalities, ambiguities around environment and task structures, and missing components in current foundation models and decision making paradigms (Section~\ref{sec:challenge}). We hope that this manuscript can serve as a stepping stone toward developing autonomous agents with next-level intelligence and more sophisticated capabilities. 

\begin{acks}
We thank Bo Dai and Douglas Eck for reviwing this manuscript.
\end{acks}


\bibliography{main}

\bibliographystyle{ACM-Reference-Format}



\end{document}